\documentclass{article}

\usepackage[main, preprint]{neurips_2026}

\usepackage[utf8]{inputenc} % allow utf-8 input
\usepackage[T1]{fontenc}    % use 8-bit T1 fonts
\usepackage{hyperref}       % hyperlinks
\usepackage{url}            % simple URL typesetting
\usepackage{booktabs}       % professional-quality tables
\usepackage{amsfonts}       % blackboard math symbols
\usepackage{nicefrac}       % compact symbols for 1/2, etc.
\usepackage{microtype}      % microtypography

\usepackage{graphicx}
\usepackage{subcaption}
\usepackage{enumitem}

\usepackage{multirow}
\usepackage{makecell}
\usepackage[table]{xcolor}

\usepackage{fontawesome5}
\usepackage{amsmath}
\usepackage{colortbl}
\usepackage{pgf}

\usepackage{xcolor}
\usepackage{pifont}
\newcommand{\cmark}{\textcolor{teal!70!black}{\ding{51}}}   % soft green/teal
\newcommand{\xmark}{\textcolor{red!70!black}{\ding{55}}}       
\usepackage{tcolorbox}
\tcbuselibrary{breakable}
\tcbuselibrary{skins}

\usepackage{tabularx}

\NewDocumentCommand{\bx}
{ mO{} }{\textcolor{teal}{\textsuperscript{\textit{bingxuan}}\textsf{\textbf{\small[#1]}}}}

\NewDocumentCommand{\zhenhailong}
{ mO{} }{\textcolor{blue}{\textsuperscript{\textit{zhenhailong}}\textsf{\textbf{\small[#1]}}}}

\NewDocumentCommand{\jiateng}
{ mO{} }{\textcolor{purple}{\textsuperscript{\textit{jiateng}}\textsf{\textbf{\small[#1]}}}}

\NewDocumentCommand{\cheng}
{ mO{} }{\textcolor{orange}{\textsuperscript{\textit{cheng}}\textsf{\textbf{\small[#1]}}}}
\NewDocumentCommand{\kunlun}
{ mO{} }{\textcolor{yellow}{\textsuperscript{\textit{kunlun}}\textsf{\textbf{\small[#1]}}}}

\NewDocumentCommand{\jeongh}
{ mO{} }{\textcolor{brown}{\textsuperscript{\textit{Jeonghwan}}\textsf{\textbf{\small[#1]}}}}

\NewDocumentCommand{\heng}
{ mO{} }{\textcolor{red}{\textsuperscript{\textit{Heng}}\textsf{\textbf{\small[#1]}}}}

\usepackage{wrapfig}
\usepackage{titletoc}

\newcommand{\ModelName}{\textit{\textcolor{darkgray}
{\textbf{MeMento}}}}{}
\newcommand{\BenchName}{\textcolor{darkgray}{$\mathit{DunphyBench}$}}{}

\title{Long-Horizon Embodied Decision-Making \\ via Multimodal Memory Compression}

\author{Bingxuan Li\textsuperscript{$1$}, Rui Yang\textsuperscript{$1$}, Cheng Qian\textsuperscript{$1$}, Jiateng Liu\textsuperscript{$1$}, Jeonghwan Kim\textsuperscript{$1$}, \\ \bfseries\ Zhenhailong Wang\textsuperscript{$1$}
\bfseries\ Manling Li\textsuperscript{$2$}, 
\bfseries\ Tong Zhang\textsuperscript{$1$},
\bfseries\ Heng Ji\textsuperscript{$1$} \vspace{2.5mm} \\
\textsuperscript{$1$}University of Illinois Urbana-Champaign, \textsuperscript{$2$}Northwestern University
}
\begin{document}

\maketitle

\begin{abstract}
Agents are increasingly expected to act not only as task executors, but also as decision-makers on behalf of human users. This shift requires agents to accumulate evidence over long horizons, interpret implicit user preferences, and compare multiple candidates under partial observations. In this work, we propose \BenchName{}, a new benchmark for evaluating agents on long-horizon human-centered embodied decision-making, where the agent must navigate through multiple embodied housing environments and make decisions that align with multi-dimensional human preferences. Unlike standard embodied reasoning tasks that often focus on procedural planning or immediate goal completion, our setting requires agents to integrate multimodal, multi-source input into coherent knowledge that supports complex reasoning across long horizon. The evaluation results reveal that there is a substantial gap between current agents and human performance. Furthermore, our diagnosis of state-of-the-art VLM-driven agents reveals that memory management is one of the bottlenecks, where raw multimodal history introduces noise that hinders decision quality. Motivated by this finding, we design \ModelName{}, a preference-conditioned multimodal memory compressor that selectively compresses decision-relevant information from long-horizon history based on user preferences with a fixed set of memory tokens. Experiments show that \ModelName{} helps VLM-driven agents improve accuracy by 7.18\%, while reducing memory usage by 85.38\% compared to the strongest baseline. 

\end{abstract}

% \heng{try to jump to the key points at the very beginning to reflect the strength of your technical approaches and innovations. For decision making in the real world, we often need to 1. remotely nagivate some environments; 2. information in text is often not enough, so it requires cross-media knowledge linking and aggregation; 3. information from multiple sources could be massive and overwhelming, given time and ememory constraints, we need methods to compress memory so the most useful information can be utilized efficiently. Then map each of these points to your solution. }

\section{Introduction}
Over long decision-making processes in the real world, information often accumulates faster than humans can track, leading to decision fatigue and degraded judgment \citep{decision_fatigue}. Consider the case of apartment searching: Imagine you are relocating to a new town and need to find a new home. You compare many homes against your preferences while repeatedly taking virtual tours. Each visit adds new visual details, spatial layouts, amenities, and trade-offs that must be remembered and compared. As the search continues, fatigue sets in, and you may eventually settle for an option that is merely ``good enough'' rather than the one that best matches their needs. This highlights the need for an automated decision partner: An intelligent assistant that can maintain long-horizon memory, compare alternatives systematically, and make decisions on behalf of users deliberately.

% \begin{figure}
%     \centering
%     \includegraphics[width=1\linewidth]{/figs/task.pdf}
%     \caption{Task illustration of \BenchName{}}
%     \label{Task}
% \end{figure}

Building such an agent to support human decision-making requires a multidimensional set of capabilities. Specifically, the agent must: (1) autonomously navigate and inspect remote environments; (2) link and aggregate information across modalities, since textual descriptions alone are often insufficient; (3) manage large volumes of accumulated evidence over long horizons under limited time and memory budgets; and (4) infer and align with implicit user preferences that may not be explicitly stated. Recent advances in large language models (LLMs) and vision-language models (VLMs) have enabled agents that can perceive, navigate, and act in interactive environments. These agents are increasingly envisioned as personal assistants that operate on behalf of users over extended interactions \citep{hong2025embodied,li2025metal,liu2026osexpertcomputeruseagentslearning, li2025echofoley, li2026pearl}. However, it remains unclear whether current agents possess the capabilities required for such long-horizon, preference-driven embodied decision-making:
\begin{center}
\vspace{-0.1in}
\textbf{Can agents make preference-aligned decisions over long-horizon embodied interactions?}
\vspace{-0.1in}
\end{center}
We consider two core challenges in this setting: preference under-specification and long-horizon grounding. Human preferences are rarely provided as complete checklists. They may be explicit, such as ``I need three bedrooms,'' but they are often implicit, indirect, or persona-driven, such as ``I often cook for friends'' or ``I prefer a calm place to work.'' To make aligned decisions, the agent must infer what these signals imply, ground them in visual observations in the embodied environments, and compare evidence across multiple candidate environments. Hence, we propose a \textbf{new embodied reasoning problem} that requires more than recognizing objects or completing navigation goals, but connecting personalized preference signals to accumulated embodied exploration experience.

\begin{wrapfigure}{r}{0.5\textwidth}
    \centering
    \vspace{-0.1in}
    \includegraphics[width=0.5\textwidth]{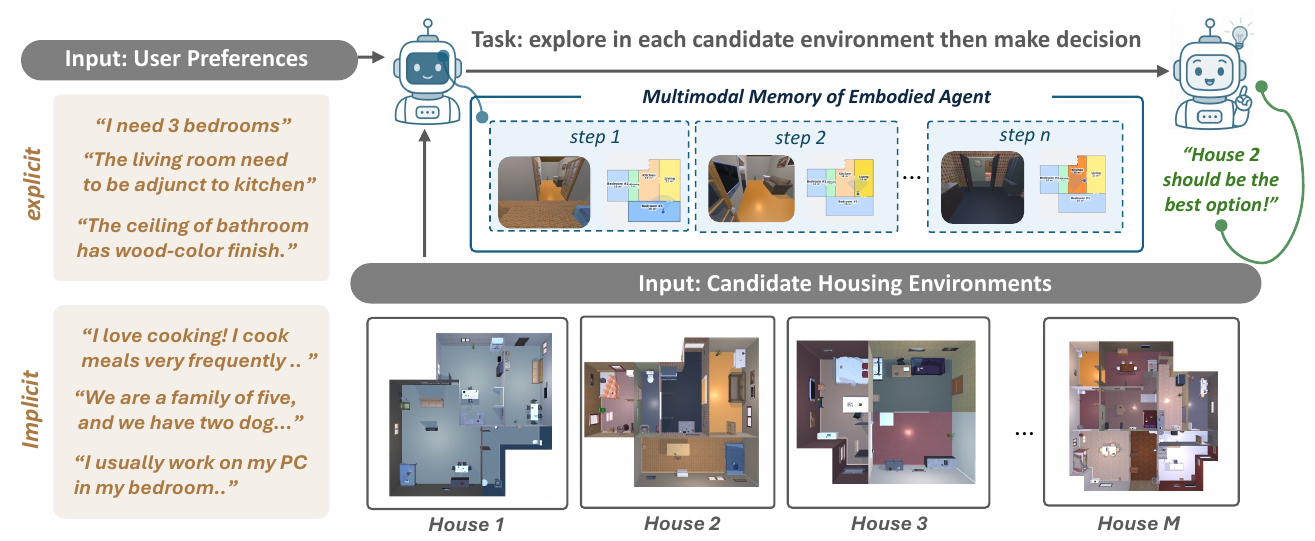}
    \caption{Task illustration of \BenchName{}.}
    \label{fig:task}
    \vspace{-0.1in}
\end{wrapfigure}

To systematically study this problem, we introduce \BenchName{}, a new benchmark to challenge agents with the task of long-horizon human-centered embodied decision-making, where the agent must navigate through embodied household environments and make decisions that align with compositional human preferences. The benchmark includes both \textbf{explicit preferences}, expressed as direct requirements, and \textbf{implicit preferences}, expressed through descriptive or persona-like signals. This design isolates whether agents can move beyond requirement checking toward preference inference and grounding. As illustrated in Table \ref{tab:capability_comparison}, unlike short-horizon benchmarks that evaluate whether an agent can complete a predefined task, \BenchName{} requires agents to gather observations across multiple environments, remember and compare evidence over time, and make a final decision.

Our evaluation results reveal a substantial gap between current agents and human performance, with the best model achieving 58.3\% accuracy versus 83.3\% for humans, and uncovers three key challenges in long-horizon embodied decision-making. First, performance degrades sharply with task complexity: accuracy drops by over 20 points from Easy to Hard settings and further collapses as the number of requirements and distractors increases, indicating strong sensitivity to compositional reasoning and cluttered decision spaces. Second, implicit preference reasoning remains a critical bottleneck, with performance consistently lagging far behind explicit settings (e.g., 49.1\% vs. 26.7\% in Easy), highlighting the difficulty of inferring underspecified user intent. Third, memory emerges as one of driver of performance for some agents: incorporating multimodal memory nearly doubles accuracy compared to memory-less agents, while tools provide only incremental gains. Our diagnosis finds that \textbf{long-horizon memory management is one of the bottlenecks}: Although adding history initially improves performance, excessive, unfiltered context introduces noise and leads to degradation, revealing that effective long-horizon multimodal memory management rather than scaling context alone is the key for building robust decision-making agents.

Motivated by this finding, we propose \ModelName{}, a preference-conditioned multimodal memory compressor that selectively compresses decision-relevant information from long-horizon history based on user preferences with a fixed set of memory tokens. Experiments show that \ModelName{} achieve 15.74\% relative improvement in accuracy (7.18\% absolute improvement), while reducing memory usage by 85.38\% compared to the strongest baseline on \BenchName{}. This result shows that \textbf{\ModelName{} can serve as an improved baseline on \BenchName{}}. Furthermore, ablation study and domain transfer experiments results show that \ModelName{}’s effectiveness stems from the synergy of preference-aware representation, multimodal memory, and hierarchical compression, and that these design choices generalize beyond \BenchName{} to also improve decision-making in web tasks.

% \heng{this method needs to be expanded; if you need space the previous paragraph can be shortened}

% \input{content/02-related_works}
\section{Preliminaries}

% \begin{table}[t]
% \centering
% \small
% \setlength{\tabcolsep}{2.5pt}
% \renewcommand{\arraystretch}{0.9}

% \begin{tabular}{lccccc}
% \toprule
% \textbf{} 
% & Spatial 
% & Visual
% & Long-Term 
% & Information 
% & Preference \\
% & Reasoning 
% & Reasoning 
% & Memory 
% & Aggregation 
% & Reasoning \\
% \midrule

% ALFWorld\citep{ALFWorld}               & \xmark & \xmark & \cmark & \xmark & \xmark \\
% ALFRED~\citep{shridhar2020alfred}       & \cmark & \cmark & \cmark & \xmark & \xmark \\

% % Behavior1K~\citep{li2023behavior}             & \cmark & \cmark & \cmark & \xmark & \xmark \\
% % EmbodiedQA~\citep{das2018embodied}             & \cmark & \cmark & \xmark & \cmark & \xmark \\
% EmbodiedBench~\citep{yang2025embodiedbench}          & \cmark & \cmark & \xmark & \xmark & \xmark \\
% EmbodiedAgentInterface~\citep{li2024embodied} & \cmark & \cmark & \cmark & \xmark & \xmark \\
% WebArena~\citep{zhou2023webarena}               & \xmark & \xmark & \cmark & \cmark & \cmark \\
% Embodied Web Agent~\citep{hong2025embodied}     & \cmark & \cmark & \xmark & \cmark & \xmark  \\
% BabyAI~\citep{chevalier2018babyai}                 & \cmark & \cmark & \xmark & \cmark & \xmark \\
% UserBench~\citep{qian2025userbench}       & \xmark & \xmark & \xmark & \xmark & \cmark  \\

% \midrule
% \textbf{\BenchName{}} 
% & \textbf{\cmark} 
% & \textbf{\cmark} 
% & \textbf{\cmark} 
% & \textbf{\cmark} 
% & \textbf{\cmark} \\

% \bottomrule
% \end{tabular}
% \vspace{0.05in}
% \label{tab:capability_comparison}
% \caption{Measured agent capability comparison across benchmarks.}
% \vspace{-0.3in}
% \end{table}

\begin{table}[t]
\centering
\scriptsize
\setlength{\tabcolsep}{6pt}
\renewcommand{\arraystretch}{0.65}

\begin{tabular}{lccccc}
\toprule
\textbf{} 
& Embodied
& Long-Term 
& Information 
& Preference
& Task Horizon / \\
& Reasoning 
& Memory 
& Aggregation 
& Reasoning 
& Max Allowed Steps \\
\midrule

BabyAI~\citep{chevalier2018babyai}                 
& \cmark & \xmark & \cmark  & \xmark 
& --- \\

ALFRED~\citep{shridhar2020alfred}       
& \cmark & \cmark & \xmark & \xmark 
& $\sim50$ \\

ALFWorld\citep{ALFWorld}               
& \xmark & \cmark & \xmark & \xmark 
& $50$ \\

% Behavior1K~\citep{li2023behavior}       
% & \cmark & \cmark & \xmark & \xmark 
% & 6--16+ primitive steps \\

% EmbodiedQA~\citep{das2018embodied}             
% & \cmark & \xmark & \cmark & \xmark 
% & task-dependent \\

WebArena~\citep{zhou2023webarena}               
& \xmark & \xmark  & \cmark & \cmark 
& $30$ \\

EmbodiedAgentInterface~\citep{li2024embodied} 
& \cmark & \cmark & \xmark & \xmark 
& $\sim40 -60$ \\

3DMem-Bench~\citep{hu20253dllm}   
& \cmark & \cmark & \xmark & \xmark 
& $24$ \\

EmbodiedBench~\citep{yang2025embodiedbench}          
& \cmark & \xmark & \xmark & \xmark 
& $15-30$ \\

Embodied Web Agent~\citep{hong2025embodied}     
& \cmark & \xmark & \cmark & \xmark  
& $\sim50$ \\

UserBench~\citep{qian2025userbench}       
& \xmark & \xmark & \xmark & \cmark  
& $20$ \\

\midrule
\textbf{\BenchName{}} 
& \textbf{\cmark} 
& \textbf{\cmark} 
& \textbf{\cmark} 
& \textbf{\cmark} 
& \textbf{$1500$} \\

\bottomrule
\end{tabular}

\vspace{0.05in}
\caption{Measured agent capability and task horizon comparison across benchmarks.}
\label{tab:capability_comparison}
\vspace{-0.35in}
\end{table}

\noindent\textbf{Task Formulation.}
We instantiate our benchmark in the house selection task as a concrete use case for long-horizon human-centered embodied decision-making. Let $N$ denote the number of user preferences and $M$ denote the number of distractors per difficulty tier. Formally, each instance consists of a preference set of size $N$, organized into seven semantic categories, and a candidate pool of size $1 + 3M$. Among these candidates, one is the \textbf{golden environment}, which fully aligns with all preferences, while the remaining $3M$ candidates are \textbf{distractors} that only partially align with them. The distractors are partitioned into three tiers (easy, medium, and hard), each containing $M$ environments. Under this setup, the model must identify the unique environment that best satisfies the full set of preferences. 

We note that the proposed setting is not specific to housing hunting task. Many real-world decision tasks, such as disaster-response planning, facility inspection, and assistive navigation, similarly require agents to aggregate multimodal observations over time and select among candidates according to human preferences and constraints.

% \heng{add a brief summary of how ProcTHOR works}

\noindent\textbf{Environment Representation.}
To ensure that evaluation is objective and free from annotation ambiguity, all preferences must be grounded in a representation that supports automatic verification. We therefore represent each environment using a structured feature vector derived from ProcTHOR\citep{procthor} (a collection of embodied housing environments with diverse floor plans). This representation captures structural properties (room counts, layout), per-room attributes (area, windows, materials), object inventories, spatial connectivity, and aesthetic features. From this representation, we define a set of deterministic predicates, which serve as the interface between natural language preferences and environment verification.

% \heng{You can say something like you just use environment selection as a use case study, but the proposed method should be applicable to many similar tasks such as disaster management...}

\noindent\textbf{Preference Formulation.}
Human decision-making involves both explicit reasoning over stated constraints and implicit inference from natural preferences. To disentangle these two capabilities, we design two independent evaluation tracks.
(1) \textbf{Explicit Track.} Preferences are directly specified as structured constraints (e.g., at least three bedrooms''), each mapping one-to-one to a verifiable predicate. (2) \textbf{Implicit Track.} Preferences are expressed as natural user statements (e.g., I cook large meals daily''), requiring the model to infer the underlying structural needs.
Each track is generated and evaluated independently, with separate golden environments and distractor pools.

\section{Evaluating Long-Horizon Human-Centered Embodied Decision-Making}

% comparison with other benchmarks tbl

\subsection{Preference Taxonomy}
% \heng{It's better to put in a real data example. I think the examples in your slides are good.}

To ensure comprehensive coverage of environment properties and prevent models from overfitting to a narrow subset of features, we organize preferences into seven semantic categories: (1) room counts and types, (2) room dimensions and area, (3) windows and lighting, (4) object presence, (5) object counts, (6) spatial connectivity, and (7) materials and finishes.
In the explicit track, each preference directly instantiates a predicate. In the implicit track, each preference is expressed as a short natural-language statement (within 20 words). See Appendix \ref{app:taxonomy} for more details and examples.

% \heng{are any of these datasets from real data? if so emphasize your contribution to creating real-world datasets}

\subsection{Data Generation}
We design pipeline for diverse and verifiable dataset construction. See Appendix \ref{app:data_generation} for details.

% \begin{figure}[thbp]
%     \centering
%     \includegraphics[width=1\linewidth]{figures/data_pipeline.png}
%     \caption{Data Construction Pipeline}
%     \label{fig:data}
%     \vspace{-0.25 in}
% \end{figure}

\noindent\textbf{Step 1. Feature Extraction.}
We convert each ProcTHOR environment ~\citep{procthor} into a structured feature vector capturing global layout, room attributes, object inventory, and spatial adjacency. A fixed set of deterministic predicates is computed from this representation, enabling fully automatic and noise-free preference verification.

\noindent\textbf{Step 2. Golden Sampling.}
We select golden environments with sufficient complexity (e.g., $\geq$ 5 rooms) that support all preference categories. To ensure diversity, we avoid similar layouts and assign a unique golden environment to each instance.

\noindent\textbf{Step 3. Preference Generation.}
For the \textbf{explicit track}, preferences are generated from features via an LLM and mapped to verifiable predicates, discarding invalid cases. For the \textbf{implicit track}, we sample satisfied predicates first, then express them as natural-language preferences without revealing the underlying constraints.

\noindent\textbf{Step 4. Distractor Construction.}
We control difficulty by grouping distractors based on preference satisfaction $S$: easy ($<0.3N$), medium($0.3N$–$0.6N$), and hard ($0.6N$–$0.9N$). Near-satisfying cases ($\geq 90\%$) are excluded. Hard distractors are further constrained to fail on different preferences.

\noindent\textbf{Step 5. Quality Control.}
We enforce that the golden satisfies all preferences, no distractor does, each tier has exactly $M$ samples, and preferences are balanced. Cross-instance redundancy is reduced via Jaccard similarity filtering. For implicit cases, each statement must uniquely imply a single predicate.
\section{Evaluation Setting and Results}
\label{sec:results}

\subsection{Evaluation Setup}

\noindent\textbf{Backbone Models.}
We evaluate four representative models with strong multimodal reasoning capability: \texttt{Qwen3VL-8B-Instruct}, \texttt{Qwen3VL-30B-Thinking}, \texttt{Gemma-4-26B-A4B-it}, and \texttt{GPT-5.4}.

\noindent\textbf{Agent Design.} To enable controlled analysis of how memory and agent configuration affect decision quality, we consider two primary variants for evaluated agent design:
\begin{itemize}[leftmargin=*, noitemsep, topsep=0pt]
    \item \textit{Memoryless Agent.} Performs one-shot reasoning over all inputs without maintaining memory.
    
    \item \textit{In-Context Memory Agent.} Maintains an external recent memory, and append to the prompt during inference. We instantiate two forms of memory: (1) \textit{Text Memory.} Stores structured textual summaries of experience (e.g., observations of recent steps). (2) \textit{Multimodal Memory.} Extends text memory with visual context (e.g. ego-view and map-view images of recent steps).
\end{itemize}

\noindent\textbf{Tools.}
We additionally provided a set of tools that can be an option that enabled during inference:

\begin{itemize}[leftmargin=*, noitemsep, topsep=0pt]
    \item \textit{Sketchpad.} A persistent workspace for recording intermediate thoughts and observations in text.
    \item \textit{Retriever.} A search tool to search relevant memory based on the agent input query.    
\end{itemize}

\noindent\textbf{Evaluation Metrics} We design three metrics to capture decision correctness, preference alignment, and efficiency:

\begin{itemize}[leftmargin=*, noitemsep, topsep=0pt]
    \item \textit{Accuracy (Acc):} We measure the fraction of instances where the agent make the correct final decision.

     \item \textit{Normalized Satisfaction Ratio (nSAT):}
     We measure partial correctness by selected satisfaction over gold requirements: 
    $
    \text{nSAT} = \frac{|R_{\text{chosen}} \cap R_{\text{gold}}|}{|R_{\text{gold}}|}
    $

    \item \textit{Exploration Efficiency (EE):}
    We measure the balance between efficiency and decision accuracy. Let $K$ denote the number of exploration steps taken, and $K_{\max}$ the maximum allowed exploration budget. We define:
   $ \text{EE} = ( 1 - \frac{K}{K_{\max}}) \times$ nSAT

\end{itemize}

Default \BenchName{} configuration is $N=7, M=19$. Human annotators are evaluated with the same evaluation protocol as agents. For more setup details, see Appendix ~\ref{app:benchmark}.

\subsection{Evaluation Results}

\subsubsection{Main Evaluation Results}

% % ===== Avg nSAT: CELL COLOR =====
% \newcommand{\avgnsatcell}[1]{%
% \pgfmathsetmacro{\normraw}{100*(#1-\minnsat)/(\maxnsat-\minnsat)}%
% \pgfmathsetmacro{\norm}{max(0,min(100,\normraw))}%
% \pgfmathsetmacro{\inv}{100-\norm}%
% \edef\temp{\noexpand\cellcolor{red!\inv!green!20!white}}%
% \temp #1\%
% }

% ===== Icons =====
\newcommand{\nomem}{\faTimes}
\newcommand{\textmem}{\faFont}
\newcommand{\mmem}{\faImage}

\newcommand{\notool}{\faTimes}
\newcommand{\sketch}{\faPen}
\newcommand{\retriever}{\faSearch}
\newcommand{\bothtool}{\faPen{\scriptsize +}\faSearch}

% ===== Normalization =====
\newcommand{\minacc}{5.6}
\newcommand{\maxacc}{83.3}
\newcommand{\minnsat}{28.9}
\newcommand{\maxnsat}{94.4}

% ===== Avg Acc: TEXT COLOR =====
\newcommand{\avgacctext}[1]{%
\pgfmathsetmacro{\normraw}{100*(#1-\minacc)/(\maxacc-\minacc)}%
\pgfmathsetmacro{\norm}{max(0,min(100,\normraw))}%
\pgfmathtruncatemacro{\invint}{100-\norm}%
\edef\acccolor{red!\invint!green!70!black}%
{\bfseries\expandafter\color\expandafter{\acccolor}#1\%}%
}

% ===== Avg SAT: TEXT COLOR =====
\newcommand{\avgnsatcell}[1]{%
\pgfmathsetmacro{\normraw}{100*(#1-\minnsat)/(\maxnsat-\minnsat)}%
\pgfmathsetmacro{\norm}{max(0,min(100,\normraw))}%
\pgfmathtruncatemacro{\invint}{100-\norm}%
\edef\acccolor{red!\invint!green!70!black}%
{\bfseries\expandafter\color\expandafter{\acccolor}#1\%}%
}

% ===== Avg Steps: BLACK TEXT ONLY =====
\newcommand{\avgstepstext}[1]{\texttt{#1}}

\begin{table*}[t]
\centering
\scriptsize
\setlength{\tabcolsep}{6.5pt}
\renewcommand{\arraystretch}{0.55}

\begin{tabular}{lcc|cc|cc|ccc}
\toprule

& \multicolumn{2}{c|}{\textbf{Agent Configuration}}
& \multicolumn{2}{c|}{\textbf{Explicit}}
& \multicolumn{2}{c|}{\textbf{Implicit}}
& \multicolumn{3}{c}{\textbf{Average}} \\

\cmidrule(lr){2-3}
\cmidrule(lr){4-5}
\cmidrule(lr){6-7}
\cmidrule(lr){8-10}

& \textit{Memory} & \textit{Tool}
& \textit{Acc.} & \textit{nSAT.}
& \textit{Acc.} & \textit{nSAT.}
& \textit{Acc.} & \textit{nSAT.} & \textit{Steps} \\

\midrule

\multirow{7}{*}{\texttt{Qwen3VL-8B-Instruct}}
 & \nomem   & \notool    & 11\% & 30\% & 6\%  & 28\% & \avgacctext{8.3}  & \avgnsatcell{28.9} & \avgstepstext{669.3} \\
& \nomem   & \sketch    & 11\% & 35\% & 11\% & 31\% & \avgacctext{11.1} & \avgnsatcell{32.8} & \avgstepstext{850.8} \\
& \textmem & \notool    & 6\%  & 31\% & 22\% & 41\% & \avgacctext{13.9} & \avgnsatcell{36.2} & \avgstepstext{946.2} \\
& \textmem & \sketch   & 39\% & 54\% & 11\% & 32\% & \avgacctext{25.0} & \avgnsatcell{43.2} & \avgstepstext{721.6} \\
& \textmem & \retriever & 11\% & 34\% & 0\%  & 25\% & \avgacctext{5.6}  & \avgnsatcell{29.3} & \avgstepstext{840.3} \\
& \mmem    & \notool    & 22\% & 40\% & 11\% & 31\% & \avgacctext{16.7} & \avgnsatcell{35.8} & \avgstepstext{560.3} \\
& \mmem    & \bothtool  & 11\% & 33\% & 22\% & 42\% & \avgacctext{16.7} & \avgnsatcell{37.4} & \avgstepstext{837.5} \\

\addlinespace[2pt]

\multirow{7}{*}{\texttt{Qwen3VL-30B-Thinking}}
& \nomem   & \notool    & 17\% & 37\% & 11\% & 33\% & \avgacctext{13.9} & \avgnsatcell{34.8} & \avgstepstext{433.3} \\
& \nomem   & \sketch    & 28\% & 45\% & 17\% & 36\% & \avgacctext{22.2} & \avgnsatcell{40.5} & \avgstepstext{313.7} \\
& \textmem & \notool    & 28\% & 43\% & 11\% & 32\% & \avgacctext{19.4} & \avgnsatcell{37.5} & \avgstepstext{345.2} \\
& \textmem & \sketch    & 28\% & 42\% & 0\%  & 24\% & \avgacctext{13.9} & \avgnsatcell{33.2} & \avgstepstext{318.3} \\
& \textmem & \retriever & 22\% & 40\% & 0\%  & 26\% & \avgacctext{11.1} & \avgnsatcell{33.0} & \avgstepstext{187.2} \\
& \mmem    & \notool    & 39\% & 51\% & 33\% & 37\% & \avgacctext{27.8} & \avgnsatcell{43.9} & \avgstepstext{336.8} \\
& \mmem    & \bothtool  & 33\% & 48\% & 28\% & 44\% & \avgacctext{35.6} & \avgnsatcell{45.9} & \avgstepstext{479.9} \\

\addlinespace[2pt]

\multirow{7}{*}{\texttt{Gemma-4-26B-A4B-it}}
& \nomem   & \notool    & 6\%  & 29\% & 6\%  & 25\% & \avgacctext{5.6}  & \avgnsatcell{29.2} & \avgstepstext{921.5} \\
& \nomem   & \sketch    & 17\% & 38\% & 28\% & 44\% & \avgacctext{22.2} & \avgnsatcell{40.8} & \avgstepstext{977.8} \\
& \textmem & \notool    & 39\% & 52\% & 33\% & 46\% & \avgacctext{36.1} & \avgnsatcell{49.3} & \avgstepstext{751.2} \\
& \textmem & \sketch    & 11\% & 33\% & 22\% & 41\% & \avgacctext{16.7} & \avgnsatcell{36.6} & \avgstepstext{1015.0} \\
& \textmem & \retriever & 39\% & 52\% & 11\% & 32\% & \avgacctext{25.0} & \avgnsatcell{41.7} & \avgstepstext{1187.2} \\
& \mmem    & \notool    & 22\% & 41\% & 22\% & 40\% & \avgacctext{22.2} & \avgnsatcell{40.0} & \avgstepstext{929.3} \\
& \mmem    & \bothtool  & 39\% & 58\% & 33\% & 51\% & \avgacctext{36.1} & \avgnsatcell{54.6} & \avgstepstext{1276.0} \\

\addlinespace[2pt]

\multirow{7}{*}{\texttt{GPT-5.4}}
& \nomem   & \notool    & 33\% & 47\% & 17\% & 35\% & \avgacctext{25.0} & \avgnsatcell{41.1} & \avgstepstext{584.1} \\
& \nomem   & \sketch    & 50\% & 58\% & 33\% & 46\% & \avgacctext{41.7} & \avgnsatcell{51.8} & \avgstepstext{393.7} \\
& \textmem & \notool    & 61\% & 64\% & 39\% & 51\% & \avgacctext{50.0} & \avgnsatcell{57.0} & \avgstepstext{225.3} \\
& \textmem & \sketch    & 56\% & 58\% & 17\% & 35\% & \avgacctext{36.1} & \avgnsatcell{46.8} & \avgstepstext{240.6} \\
& \textmem & \retriever & 44\% & 52\% & 33\% & 46\% & \avgacctext{38.9} & \avgnsatcell{49.0} & \avgstepstext{262.9} \\
& \mmem    & \notool    & 67\% & 68\% & 50\% & 57\% & \avgacctext{58.3} & \avgnsatcell{62.6} & \avgstepstext{273.9} \\
& \mmem    & \bothtool  & 50\% & 58\% & 44\% & 52\% & \avgacctext{47.2} & \avgnsatcell{54.6} & \avgstepstext{229.2} \\

\midrule

Human 
& -- & \notool 
& 86\% & 92\%
& 81\% & 96\% 
& \avgacctext{83.3} & \avgnsatcell{94.4} & \avgstepstext{216.7} \\

\bottomrule
\end{tabular}

\caption{
Performance comparison. Agent configurations are defined by \textit{Memory} and \textit{Tool}.
Memory types include: no memory (\nomem), text-only memory (\textmem), and multimodal memory (\mmem).
Tool usage includes: no tool (\notool), sketch-based reasoning (\sketch), retrieval (\retriever), and both tools (\bothtool).
}
\label{tab:main_results}
\vspace{-0.15in}
\end{table*}

Table~\ref{tab:main_results} summarizes the performance across all configurations. Overall, agent performance remains substantially below human level, with the best agent achieving 58.3\% accuracy compared to 83.3\% for humans. We highlight the following key observations.

\begin{figure}[tbhp]
    \centering
    \begin{subfigure}[t]{0.53\linewidth}
        \centering
        \includegraphics[width=\linewidth]{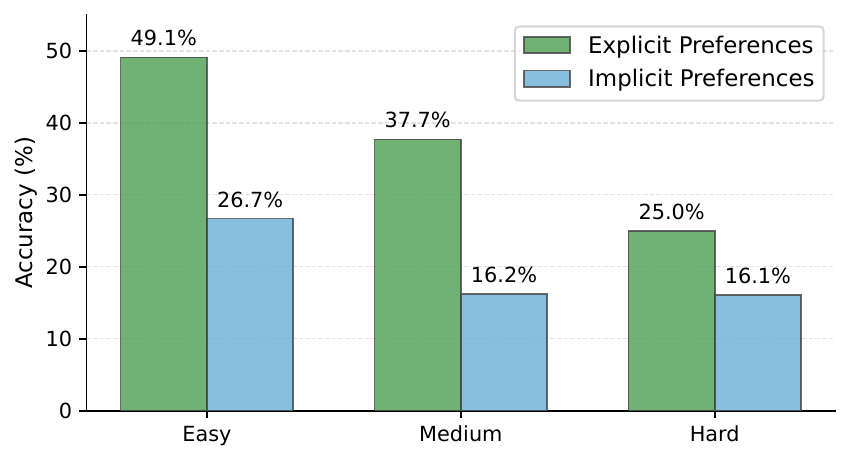}
    \end{subfigure}
    \hfill
    \begin{subfigure}[t]{0.45\linewidth}
        \centering
        \includegraphics[width=\linewidth]{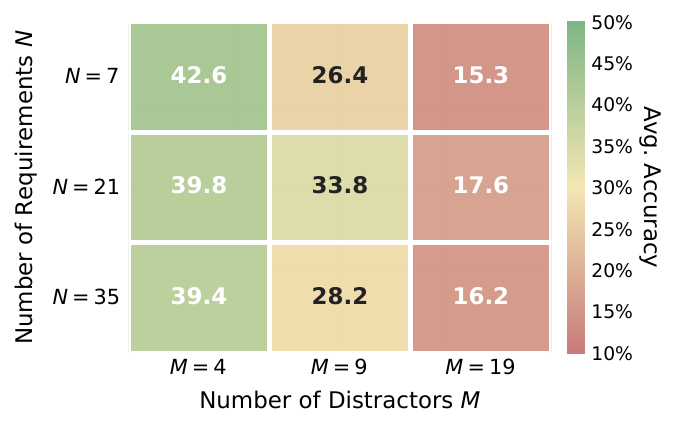}
    \end{subfigure}
    \caption{Performance of \texttt{Qwen3VL-8B-Instruct} agent under varying task settings. \textbf{Left}: Accuracy across difficulty levels. \textbf{Right}: Impact of increasing number of requirements($N$) and distractors($M$).}
    \vspace{-0.1in}
    \label{fig:broad_analysis}
\end{figure}

\begin{figure}[tbhp]
    \centering
    \begin{subfigure}[t]{0.55\linewidth}
        \centering
        \includegraphics[width=\linewidth]{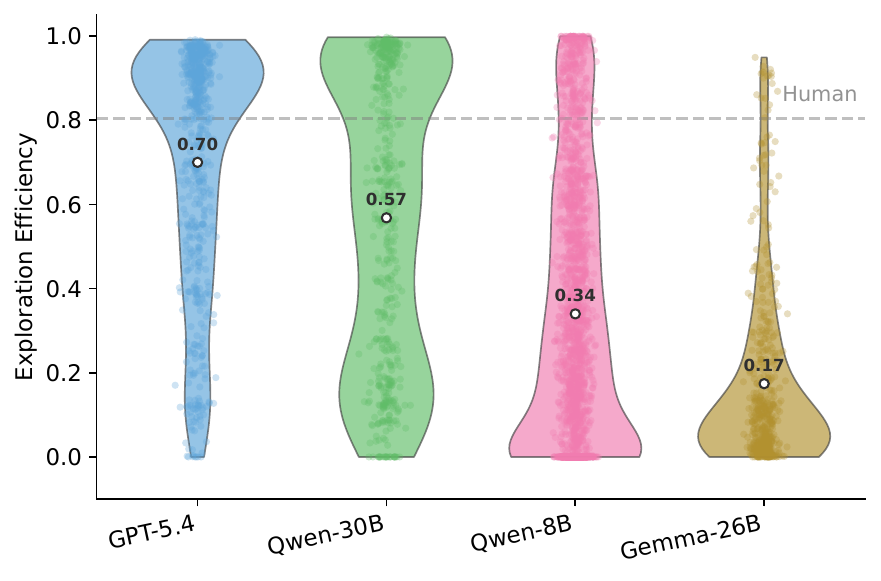}
    \end{subfigure}
    \hfill
    \begin{subfigure}[t]{0.42\linewidth}
        \centering
        \includegraphics[width=\linewidth]{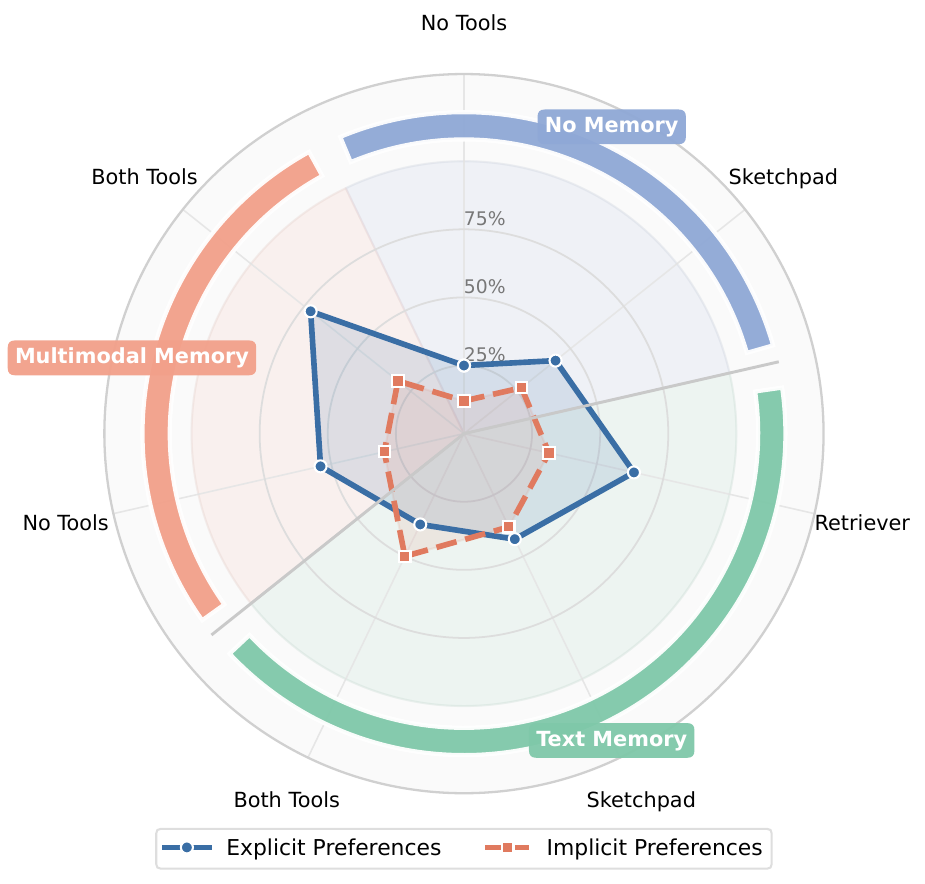}        
    \end{subfigure}
    \caption{Performance analysis on efficiency and agent configuration. \textbf{Left}: Exploration efficiency across models. \textbf{Right}: Performance under different agent (memory, tool) configurations.}
    \label{fig:depth_analysis}
    \vspace{-0.2in}
\end{figure}

\noindent\textbf{Insight 1. Performance degrades with task complexity.}
As shown in Figure~\ref{fig:broad_analysis} (Left), accuracy drops from 49.1\% to 25.0\% (explicit) and from 26.7\% to 16.1\% (implicit) when moving from Easy to Hard settings. Figure~\ref{fig:broad_analysis} (Right) further shows that increasing distractors from $M=4$ to $M=19$ reduces accuracy from 42.6\% to 15.3\%. These results suggest that agents are sensitive to compositional complexity and cluttered decision spaces.

\noindent\textbf{Insight 2. Agents struggle to interpret implicit preferences.}
As illustrated in Table~\ref{tab:main_results}, performance under implicit preferences is significantly lower than explicit ones across all setting. For example, in Figure~\ref{fig:broad_analysis} (Left), accuracy drops from 49.1\% (explicit) to 26.7\% (implicit) under Easy difficulty, and even at the best configuration, implicit performance remains significantly lower than explicit, highlighting the difficulty of inferring underspecified user intent. 

\noindent\textbf{Insight 3. Memory can be highly beneficial for performance.}
Figure~\ref{fig:depth_analysis} (Right) shows that incorporating memory leads to improvements across configurations for some models. For instance, in GPT-5.4, accuracy improves from 25.0\% (no memory) to 50.0\% (text memory) and further to 58.3\% (multimodal memory), demonstrating that memory contributes substantial improvement in total. This demonstrates that effective aggregation of long-horizon history context is critical.

\noindent\textbf{Insight 4. Tools can provide complementary but limited gains.}
While tool usage (i.e. sketchpad or retriever) improves performance for some models, the gains are modest compared to memory. However, tools provide inconsistent gains and may even hurt when they introduce additional context or reasoning overhead, hence, tools alone are insufficient to overcome the long-horizon challenge.

\noindent\textbf{Insight 5. Balancing efficient exploration and accuracy remains challenging for agents.}
Step efficiency alone does not guarantee better decision-making. Some agents use fewer exploration steps but still achieve low accuracy, suggesting that they may stop early without collecting enough preference-relevant evidence. As shown in Figure \ref{fig:depth_analysis} (Left), human can find a better balance between exploration cost and decision quality.

\subsubsection{Long-Horizon Multimodal Memory Management is One of Bottlenecks}
\label{sec:memory_budget}

We conduct a failure mode analysis, and find out current agents often fail due to ineffective memory management. In particular, many errors arise when agents lose track of where they are or which regions they have already explored (21\%), and becoming stuck in the same environment to revisit explored areas (13\%), indicating difficulty in tracking visited states and aggregating evidence over time. We further analyze how the memory budget affects agent performance. Increasing the amount of historical context from 0 turn to 10 turns initially improves accuracy by 17\%, indicating that additional observations help the agent make decisions. However, beyond a moderate budget (15 turns), performance begins to decline. This suggests that excessive context introduces irrelevant or distracting information.

\vspace{1pt}
\noindent
\begin{tcolorbox}[
    enhanced,
    colback=green!5,
    colframe=green!35!black,
    boxrule=0.6pt,
    left=10pt,
    right=10pt,
    top=8pt,
    bottom=8pt,
    borderline west={3pt}{0pt}{green!50!black},
]
\textbf{\textcolor{green!40!black}{Key Takeaway}}: 
\emph{Memory scaling is non-monotonic for agent performance}: While additional history can help, excessive unfiltered memory introduces noise and degrades performance.
\end{tcolorbox}
\vspace{-0.4pt}

\vspace{-0.1in}
\section{Preference-Conditioned Multimodal Memory Compression}
\label{sec:method}

As discussed in Section~\ref{sec:memory_budget}, long-horizon embodied decision-making requires accumulating extensive multimodal history, yet conditioning on the full history is inefficient and harmful, as irrelevant or redundant information overwhelms the context and degrades performance. Motivated by this, we propose to compress multimodal history into a small set of \textbf{preference-relevant memory tokens}. 

\begin{figure}[h]
    \centering
    \includegraphics[width=\linewidth]{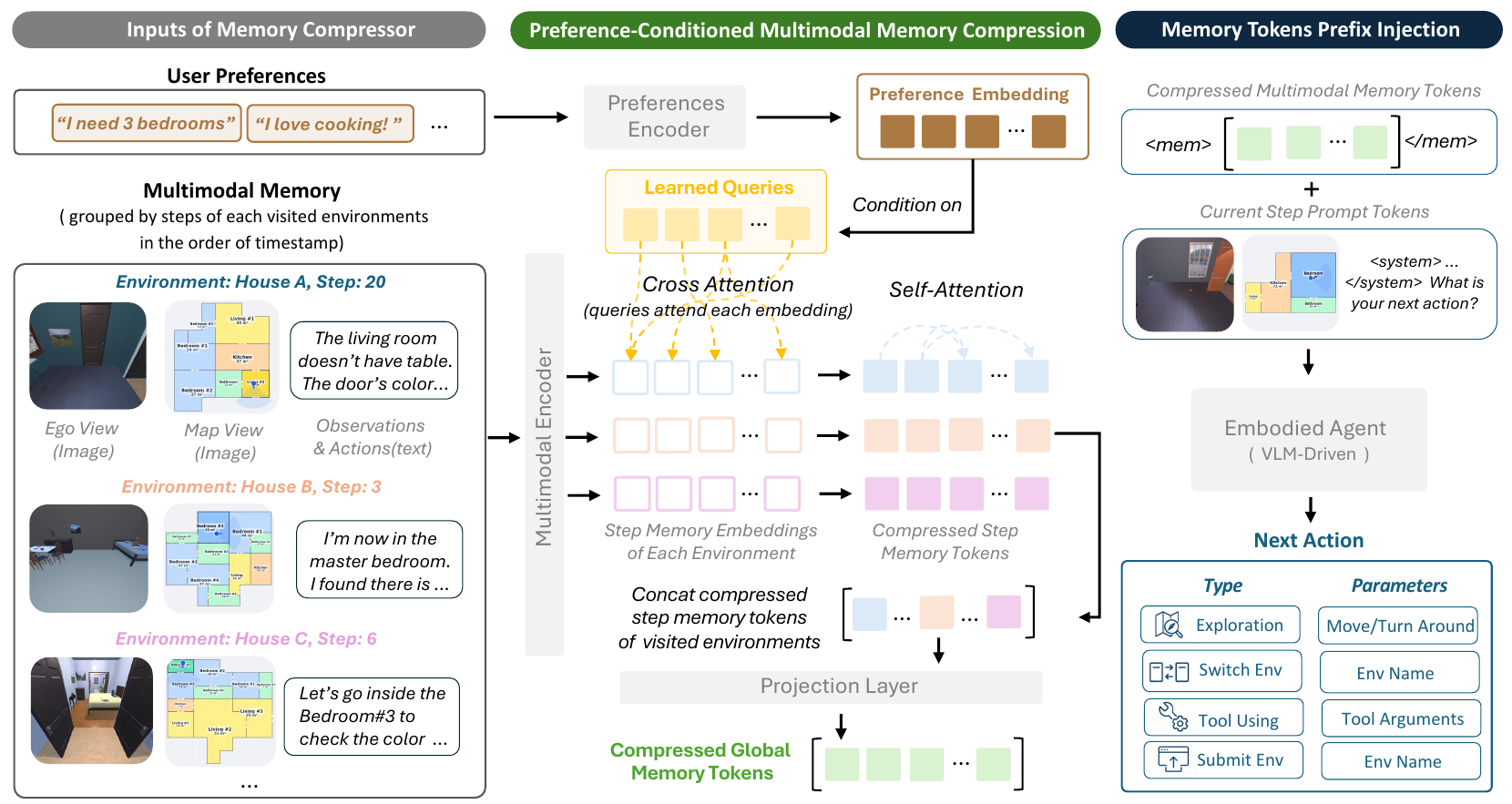}
    \caption{Overview of \ModelName{}: Preference-conditioned multimodal memory compression.}
    \label{fig:mmc}
    \vspace{-0.2in}
\end{figure}

As shown in Figure~\ref{fig:mmc}, we instantiate the \textbf{compressor as a preference-conditioned Perceiver-style module}. Our key idea is to have a fixed set of learnable queries as memory tokens, compress preference-relevant information from raw multimodal history.

% \zhenhailong{I think we should revise this figure (also need to remove the AI artifacts); currently the arrows are very confusing, we should make clear that we use a fixed set of learnable queries as memory tokens, compress info from raw multimodal history; Figure 8.a in my previous Paxion paper: https://arxiv.org/pdf/2305.10683, might be a reference?)}

\subsection{Preference and Multimodal Memory Encoding}

Given user requirements, we first construct the preference representation. The requirements may include explicit constraints, such as ``3 bedrooms,'' or implicit preferences, such as ``I cook often.'' To standardize the input, we prompt the same frozen VLM backbone to transform these requirements into keywords, and extract the text embeddings of these keywords as preference tokens. These tokens are pooled into a single preference embedding $\bar{\mathbf{z}}_r$, which serves as the signal for memory selection.

We then encode encode each explored step in multimodal memory into a multimodal step representation. Each step contains the ego-view image, map view, observation text, and action text:
\[
x_t = \{\text{ego image}, \text{map image}, \text{observation}, \text{action}\}, 
\qquad 
\mathbf{e}_t = f_{\mathrm{enc}}(x_t).
\]
To preserve temporal and environment identity, we augment each step representation with step id and environment id:
$\tilde{\mathbf{e}}_t
=
\mathbf{e}_t
+
\mathbf{p}^{\mathrm{time}}_t
+
\mathbf{p}^{\mathrm{env}}_k,$
where $\mathbf{p}^{\mathrm{time}}_t$ encodes the step index and $\mathbf{p}^{\mathrm{env}}_k$ encodes the visited environment name.

\subsection{Preference-Conditioned Memory Compression}
To compress the encoded dense step representations, we let the preference-conditioned learned queries decide which information should be retained. We compress each environment independently to prevent evidence from different candidates from being mixed.

Given a learned query set $\mathbf{Q}_0 \in \mathbb{R}^{M \times d}$, we inject the preference embedding into the queries with a projection network  $\phi(\cdot)$, resulting the preference-conditioned queries:
$\mathbf{Q}_r = \mathbf{Q}_0 + \phi(\bar{\mathbf{z}}_r)$.

For each environment $\mathcal{E}_k$, the preference-conditioned queries attend over the corresponding step representations. A self-attention layer then fuses information across steps to produce a compact set of environment-level memory tokens:
\[
\mathbf{H}^{(k)}
=
\mathrm{SelfAttn}
\left(
\mathrm{CrossAttn}
\left(
\mathbf{Q}_r,
\{\tilde{\mathbf{e}}_t : t \in \mathcal{E}_k\},
\{\tilde{\mathbf{e}}_t : t \in \mathcal{E}_k\}
\right)
\right).
\] 

The environment-level memory tokens are then concatenated together to form the compressed global memory tokens $\mathbf{H}_{\mathrm{global}}$.

\subsection{Compressed Memory Injection}

The compressed global memory tokens are projected into the frozen VLM embedding space through a trainable adapter:
$\tilde{\mathbf{M}} = \psi(\mathbf{H}_{\mathrm{global}}),$
where $\psi(\cdot)$ maps compressor outputs to the input-embedding dimension of the VLM. Then we inject the memory tokens with a structured prefix, and prepend to the current step prompt tokens:
$[\tilde{\mathbf{M}};\mathbf{X}_{\mathrm{current}}],$
where $\mathbf{X}_{\mathrm{current}}$ denotes the current visual and textual prompt. The VLM backbone remains frozen, and the compressed memory acts as a soft prompt that provides long-horizon context for embodied reasoning and next action prediction.

\subsection{Training and Inference}
\label{sec:training}

We train \ModelName{} in two stages. The first stage trains the compressor to locate preference-relevant evidence from long multimodal histories. The second stage trains the compressor and projection adapter to align the compressed multimodal memory with the VLM embedding space.

\noindent\textbf{Stage 1: Preference-Guide Compression Training.}
In this stage, we add a prediction head to the compressor and train it to predict the step-level preference-relevance score. We start with construct the step-level relevance supervision dataset: For each trajectory, each history step is annotated an relevance label indicating whether it contains information relevant to the preference. We design four relevance label types:
$y_t \in \{\text{positive}, \text{negative}, \text{unknown}, \text{irrelevant}\}.$
% Positive evidence indicates that a step supports a requirement, negative evidence indicates that a relevant region was observed but the requirement appears unsatisfied, unknown indicates that the relevant evidence was not sufficiently observed, and irrelevant indicates that the step is unrelated to the preference. This distinction is important for preferences involving absence or negation, such as ``no table in the living room.''  

Given the preference embedding $\bar{\mathbf{z}}_r$ and step representation $\tilde{\mathbf{e}}_t$, we train the classifier $s_t = g(\bar{\mathbf{z}}_r, \tilde{\mathbf{e}}_t)$ with cross-entropy loss:
$\mathcal{L}_{\mathrm{rel}}
=
\mathrm{CE}(s_t, y_t).$

We also align the compressor's cross-attention with the relevance labels. Let $\alpha_t$ denote the step-level attention score obtained by averaging cross-attention weights across heads and memory queries. We convert relevance labels into a target distribution $\tilde{y}_t$ over steps and minimize:
$\mathcal{L}_{\mathrm{attn}}
=
-\sum_t \tilde{y}_t \log(\alpha_t + \epsilon).$
The final stage 1 training loss is:
$\mathcal{L}_{\mathrm{stage1}}
=
\mathcal{L}_{\mathrm{rel}}
+
\lambda \mathcal{L}_{\mathrm{attn}}$

% where $\lambda$ controls the strength of attention alignment, where we set to 0.2 in our implementation.

\noindent\textbf{Stage 2: Alignment Training.} In this stage, we train the compressor and projection adapter to make the compressed memory readable by the frozen VLM. We construct the training dataset with preference-related QA questions, such as whether an object was observed in a room, whether a relation was absent. Each training instance consists of compressed memory tokens, a question about a previously explored environment, and the groundtruth answer( \texttt{yes}, \texttt{no}, or \texttt{unknown}). The VLM receives only the structured memory prefix and the question prompt. During this stage, the VLM backbone remains frozen. Gradients pass through the frozen VLM computation graph into the trainable adapter and the compressor. Given adapted memory tokens $\tilde{\mathbf{M}}$, a question prompt $q$, and target answer tokens $a$, we optimize the language modeling loss only on the answer tokens:
$\mathcal{L}_{\mathrm{stage2}}
=
-\sum_{i \in \mathcal{T}_{\mathrm{ans}}}
\log p_{\theta}
\left(
a_i
\mid
a_{<i}, \tilde{\mathbf{M}}, q
\right),$
where $\mathcal{T}_{\mathrm{ans}}$ denotes the target answer positions. Prompt and memory-prefix positions are ignored.

\subsection{Inference}

At inference time, the prediction head is dropped. Given a new interaction history and user preferences, \ModelName{} encodes and compresses each environment with preference-conditioned queries, then produce the global memory tokens and projects into the frozen VLM embedding space.
\subsection{Experiments}

\subsubsection{Setup}

We employ \texttt{Qwen3VL-8B-Instruct} as the base model across all experiments and compare against four representative memory baselines:
(1) \textbf{Memoryless}: no memory.
(2) \textbf{Text (Recent) Memory}: retains textual observations from the most recent 20 steps.
(3) \textbf{Multimodal (Recent) Memory}: retains multimodal history (textual observations and ego-centric images) from the most recent 20 steps.
(4) \textbf{Text Summarization Memory}: applies the same model to compress the full interaction history till the current step into a textual summarization. (5) \textbf{RAG}: enables text/image retrieval tools. The experiments is conducted on \BenchName{} with same protocol as Section \ref{sec:results}.

\subsubsection{Results}

\begin{figure}[btp]
    \centering

    \begin{subfigure}[t]{0.53\linewidth}
        \vspace{0pt}
        \centering
        \includegraphics[width=\linewidth]{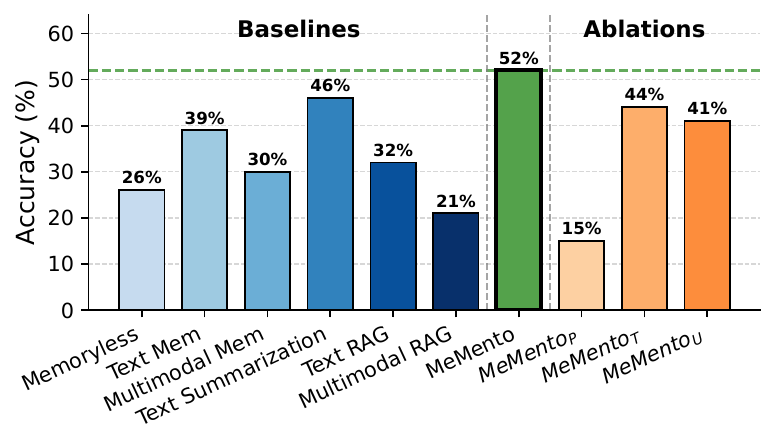}
            \vspace{-0.25in}
    \end{subfigure}
    \hfill
    \begin{subfigure}[t]{0.45\linewidth}
        \vspace{0pt}
        \centering
        {\scriptsize
\setlength{\tabcolsep}{0.9pt}
\renewcommand{\arraystretch}{1.1}
\begin{tabular}{l c}
\toprule
\textbf{Method} & \textbf{F1 ($\Delta$ to Oracle)} \\
\midrule
Text Recent Memory & 0.310 \; ($\Delta$=0.506) \\
Text Full History Summary & 0.556 \; ($\Delta$=0.260) \\
Multimodal Recent Memory & 0.333 \; ($\Delta$=0.483) \\
Retrieved Image Memory (CLIP) & 0.496 \; ($\Delta$=0.320) \\
\midrule
$\ModelName{}_{zero}$ & 0.071 \; ($\Delta$=0.745) \\
$\ModelName{}_{random}$ & 0.048 \; ($\Delta$=0.768) \\
\rowcolor{gray!20!white}
\ModelName{} & 0.719 \; (\textbf{$\Delta$=0.097}) \\
\midrule
Oracle Evidence Images (upper bound) & 0.816 \\
\bottomrule
\end{tabular}
}
    \end{subfigure}
    \caption{\ModelName{} experiment results. \textbf{Left}: Main Experiment. \textbf{Right}: Memory Probing QA}
    \label{fig:exp}
    \vspace{-0.22in}
\end{figure}

Figure~\ref{fig:exp} compares (Left) decision accuracy and memory cost across different memory strategies. Our method achieves the highest accuracy (52.78\%), outperforming all baselines while using substantially fewer memory tokens. In contrast, multimodal recent memory incurs the largest cost ($\sim$60K tokens), followed by summarization-based memory ($\sim$26K tokens) and text recent memory ($\sim$20K tokens). Our method reduces memory usage to only $\sim$3.8K tokens, achieving 85.38\% reduction compared with the strongest baseline. See Appendix \ref{app:method_exp} for more details.

\section{Discussion}

\subsection{Why does \ModelName{} work in \BenchName{}?}
We conduct an ablation study to assess the contribution of each component in \ModelName{}. Specifically, we compare three variants:
(1) $\ModelName{}_{P}$: removes the preference keywords generation module and directly encodes the requirement into a dense embedding;
(2) $\ModelName{}_{T}$: performs compression using only textual memory, excluding visual information;
(3) $\ModelName{}_{U}$: removes the hierarchical two-stage design and instead compresses the entire memory in a single stage.

\noindent\textbf{Results.} All components contribute to performance, but their impact differs. Removing preference predicates causes the largest drop (from 52.78 to 15.67), showing that explicit preference grounding is critical for alignment. Removing visual memory or hierarchical compression leads to moderate degradation (44.44 and 41.67), indicating that multimodal evidence and structured long-horizon aggregation provide complementary gains. Overall, \ModelName{} relies on the synergy of these components. See Appendix \ref{app:ablation} for more details.

\subsection{Why Multimodal Memory Compression in Our Proposed Task?}
\vspace{-0.1in}
We design a memory probing task to evaluate the advantage of multimodal memory compression in long-horizon embodied decision-making. In real-world, we often need to recall fine-grained details after completing an apartment tour when making-decision, such as \emph{``Was there a window near the sink?''} We therefore ask the agent to answer follow-up \textit{preference-related} questions about previously explored environments using only its stored memory. We compare \ModelName{} against three memory baselines: text recent memory, text summary memory, multimodal recent memory, and retrieved image. We further evaluate zero-memory and random-memory variants of \ModelName{}, and choose oracle evidence images baseline as the upper bound.

\noindent\textbf{Results.}
As shown in Table~\ref{fig:exp}, \ModelName{} achieves the best non-oracle macro-F1 score (0.719), outperforming other baselines. It also has the smallest distance to the oracle upper bound (0.097), indicating that preference-conditioned multimodal compression preserves more decision-relevant visual evidence than recent-memory or text-based compression baselines. In contrast, the zero and random memory variants perform substantially worse, with macro-F1 scores of 0.071 and 0.048, respectively, confirming that the learned memory tokens encode useful information rather than acting as a generic prompt signal. See Appendix~\ref{app:memory_qa} for more details.

\subsection{Does \ModelName{} work in other tasks?}
\vspace{-0.1in}

To investigate the generalizability of our proposed method, we further evaluate \ModelName{} on 100 web-agent tasks using the web environment from Embodied Web Agent \citep{hong2025embodied}, where agents must make decisions from multi-turn interaction. Each task contains explicit requirements, such as price or location constraints. We compare \ModelName{} with representative bassline: (1) memory-less agent, (2) recent multi-model memory agent, (3) text-summary memory agent. 
%We report task success rate and memory token usage.

\noindent\textbf{Results.}
\ModelName{} achieves the best performance with significantly higher efficiency. Memory-less agents reach 34\% success without memory, while recent multimodal memory improves performance to 69\% but requires 4.5k tokens. Text summarization further improves to 72\% with 1k tokens. In contrast, \ModelName{} achieves 83\% success using only 512 tokens, demonstrating memory efficiency and generalization to web-based decision-making. See Appendix \ref{app:web_agent} for more details.

\section{Conclusion}
\vspace{-0.1in}
We introduce \BenchName{}, a benchmark for long-horizon, human-centered embodied decision-making. Our analysis reveals that long-horizon memory management is the key bottleneck. To address this, we propose \ModelName{}, a preference-conditioned memory compression framework that distills decision-relevant information into compact representations. Experiment results show that our approach is an improved baseline that enhance both decision accuracy and memory efficiency.

\bibliographystyle{plainnat}
\bibliography{reference}

%%%%%%%%%%%%%%%%%%%%%%%%%%%%%%%%%%%%%%%%%%%%%%%%%%%%%%%%%%%%
% \appendix
% \input{content/Appendix}

%%%%%%%%%%%%%%%%%%%%%%%%%%%%%%%%%%%%%%%%%%%%%%%%%%%%%%%%%%%%

% \newpage
% \input{checklist.tex}

\end{document}